\documentclass[10pt, conference, compsocconf]{IEEEtran}

\ifCLASSOPTIONcompsoc
  \usepackage[nocompress]{cite}
\else
  \usepackage{cite}
\fi

\ifCLASSINFOpdf
  \usepackage[pdftex]{graphicx}
\else
  \usepackage[dvips]{graphicx}
\fi
\ifCLASSOPTIONcompsoc
  \usepackage[caption=false,font=footnotesize,labelfont=sf,textfont=sf]{subfig}
\else
  \usepackage[caption=false,font=footnotesize]{subfig}
\fi

\usepackage[cmex10]{amsmath}
\usepackage{url}

\usepackage{bbm}
\usepackage{hyperref} 

\begin{document}
\title{Directional Decision Lists}

\author{\IEEEauthorblockN{Marc Goessling}
\IEEEauthorblockA{Department of Statistics\\
University of Chicago\\
Chicago, IL 60637\\
Email: goessling@uchicago.edu}
\and
\IEEEauthorblockN{Shan Kang}
\IEEEauthorblockA{Robert Bosch LLC\\
Research and Technology Center North America\\
Palo Alto, CA 94304\\
Email: shan.kang@us.bosch.com}}

\maketitle

\begin{abstract}
In this paper we introduce a novel family of decision lists consisting of highly interpretable models which can be learned efficiently in a greedy manner. The defining property is that all rules are oriented in the same direction. Particular examples of this family are decision lists with monotonically decreasing (or increasing) probabilities. On simulated data we empirically confirm that the proposed model family is easier to train than general decision lists. We exemplify the practical usability of our approach by identifying problem symptoms in a manufacturing process.
\end{abstract}


\section{Introduction}
Decision lists are simple models consisting of an ordered collection of probabilistic rules. Their main advantage is that they are easy to interpret. Indeed, each model prediction is accompanied by a justification in terms of a condition that was satisfied. Decision lists are traditionally learned through greedy optimization procedures. These methods are extremely scalable and hence can be applied to very large datasets. However, they sometimes produce suboptimal models. As a remedy it was proposed to use more extensive search procedures which can overcome the drawbacks of greedy learning methods. But this comes at the price of higher computational complexity.

In this work we propose a different solution which attempts to provide accurate models while keeping the computational costs low. We try to achieve that by restricting the family of decision lists to a smaller subfamily which is more amenable to greedy learning. The basic idea is to explicitly specify the \emph{direction} of the rules. This simple concept eliminates one of the core problems of greedy selection, namely searching in the `wrong' direction.

We start with an overview of general decision lists in section \ref{sec:decision_lists}. We then illustrate a simple example for which greedy search procedures fail to provide an accurate model. Our new model family is motivated by this example and is formally introduced in section \ref{sec:directional}. We continue in section \ref{sec:experiments} with an experimental comparison between learning of directional decision lists and learning of general decision lists. We end with an experiment on real data from a manufacturing process and determine indicators for problem cases.

\section{Decision Lists}
\label{sec:decision_lists}

For binary classification ($y=0$ vs $y=1$) general decision lists \cite{rivest1987learning} are of the form:

\vspace{.5em}
\begin{quote}
IF cond$_1$ THEN $p_1$\\
ELSEIF cond$_2$ THEN $p_2$\\
$\cdots$\\
ELSEIF cond$_K$ THEN $p_K$\\
ELSE $p_\mathrm{def}$
\end{quote}
\vspace{.5em}
The conditions are chosen from a pool of binary features and $p_k \in [0,1]$ are the corresponding probabilities for $y=1$. For discrete variables the conditions are simply of the kind $\mathrm{attribute} = \mathrm{value}$. For continuous variables some form of discretization is used (e.g., $\mathrm{attribute} \in \mathrm{interval}$). It is also possible to use conjunctions of multiple primitive features. The first condition that applies to a given data point determines the predicted probability. If no condition applies then the default success probability $p_\mathrm{def}$ is used. Since a rule can only be employed if all previous conditions were $\mathrm{false}$ it means that a decision list implicitly models interactions between the selected features. Despite their simplicity decision lists are able to achieve competitive accuracy \cite{letham2015interpretable}.

Decision lists are attractive because their decision process is completely transparent to the user. The conditions can be interpreted as `reasons' for the corresponding predictions. This allows for insight into the data and makes these models more likely to be used in practice compared to black-box models which only yield predictions without providing a comprehensible justification. Decision lists are similar to decision trees \cite{breiman1984classification}, both model families create a recursive partitioning of the predictor space. However, they differ in the representation they use. Decision lists use a linear ordering of the regions while decision trees use a hierarchy. The linear structure in decision lists increases interpretability and allows for the introduction of monotonicity concepts as in \cite{Wang2015falling}.

Finding the optimal decision list (e.g., in terms of likelihoods) for a given dataset is in general an NP-hard problem. Consequently, heuristic separate-and-conquer procedures \cite{clark1989cn2,quinlan1993c4,cohen1995fast,frank1998generating} are traditionally used for learning. The computationally most efficient method is hill climbing which greedily selects the best additional rule at each step. This is an extremely scalable procedure but it sometimes gets stuck in rather poor local optima. A less greedy extension is beam search \cite{furnkranz1999separate} which at each step maintains the top $B$ candidates, called beams. For $B=1$ this method reduces to hill climbing. For larger $B$ the shortcomings of greedy selections may be alleviated. The computational cost increases linearly in $B$. An alternative approach is stochastic search \cite{letham2015interpretable,Wang2015falling} which uses an MCMC procedure to explore the whole space of decision lists. At each iteration a small modifications to the current decision list is proposed. The acceptance probability for the proposal depends on the performance of the modified decision list. Local optima can often be overcome with such a method but the computational cost is much higher than for greedy learning algorithms.

A related area of research is associative classification \cite{liu1998integrating}. The starting point for this approach is to generate a large collection of (unordered) rules by finding all class association rules which satisfy the desired support and confidence requirements. This causes a significant computational burden compared to greedy rule learning algorithms. Moreover, predictions often use an average of multiple rules \cite{li2001cmar,yin2003cpar} which leads to less intuitive classifiers compared to decision lists.

\begin{figure}[!t]
\center
\includegraphics[width=.45\textwidth]{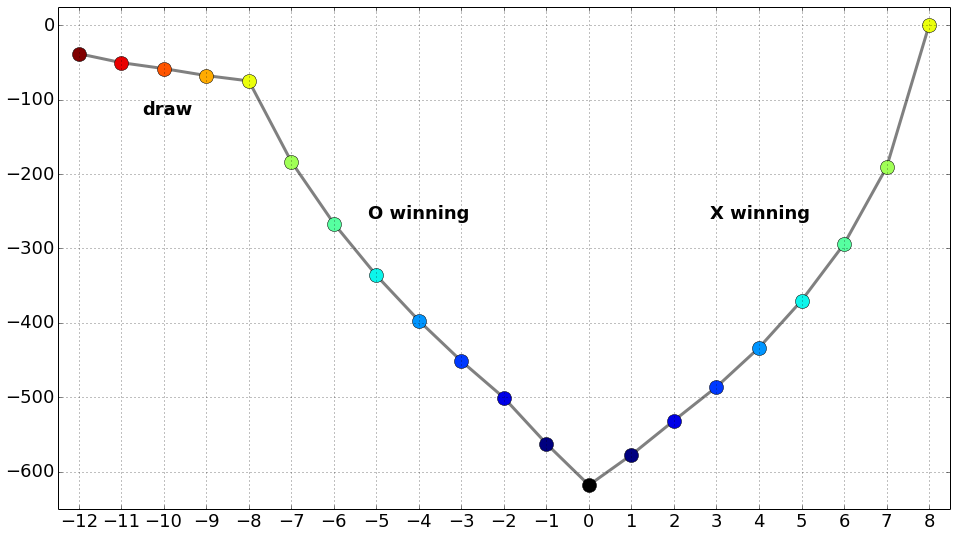}
\caption{Log-likelihood of tic-tac-toe data for decision lists of different length. The likelihood in the negative direction initially increases more rapidly.}
\label{fig:tictactoe}
\end{figure}

\section{Directional Decision Lists}
\label{sec:directional}

Greedy algorithms learn rules sequentially. At each step they can choose to select an indicative feature for $y=0$ or an indicative feature for $y=1$. Often one of the classes is easier to identify than the other (for example, characterizing spam emails may be easier than characterizing all regular emails). We observed that failure of the greedy search is frequently caused by selecting rules in the beginning which try to characterize the more complex class (which is not always the larger class). We illustrate this with a concrete example. The tic-tac-toe endgame dataset from the UCI machine learning repository\footnote{\url{http://archive.ics.uci.edu/ml}} consists of all possible board configurations at the end of tic-tac-toe games. For each cell of the 3-by-3 grid the data vectors indicate whether the cell is occupied by player X or by player O or whether it is blank. The task is to determine whether X won the game. The dataset is not symmetric in the two players because it is assumed that X started the game. Draws are counted as negative examples (i.e., X did not win the game). It is possible to obtain perfect discrimination with 8 rules corresponding to the ways in which three X markers can be placed in a horizontal, vertical or diagonal row. However, hill climbing will try to characterize the ways in which O can win and eventually gets stuck trying to characterize all draws. This happens because the purity measure (here: likelihood) for decision lists grown in the O direction initially increases faster than for decision lists grown in the X direction, see figure \ref{fig:tictactoe}. Note that a beam search would have to keep track of more than 100 beams in order to find the global optimum. This is because many different decision lists essentially point in the same direction.

The example suggests a simple fix. We can explicitly specify which of the classes will be characterized by the decision rules and ensure that the greedy search will be performed in that direction. For unordered rules the direction can simply be defined as the correlation (positive or negative) between the chosen feature and the class label \cite{liu1999pruning}. However, since decision lists are based on ordered rules an alternative definition is needed which we propose now. A directional decision list that characterizes the positive class ($y=1$) is defined as satisfying
\begin{equation}
\begin{split}
P(y=1 \,|\, \overline{\textrm{cond}_1},\ldots,\overline{\textrm{cond}_{k-1}},\textrm{cond}_k)\\
\geq P(y=1 \,|\, \overline{\textrm{cond}_1},\ldots,\overline{\textrm{cond}_k})
\label{eq:condition}
\end{split}
\end{equation}
for all $k=1,\ldots,K$. The overline denotes negation. In words this requirement means that the success probability for samples covered by rule $k$ is larger than the success probability of the remaining samples for which none of the first $k$ conditions applies. Note that decision lists recursively partition the predictor space $R$ into disjoint unions of regions $R_1 \cup R_{>1} = R_1 \cup R_2 \cup R_{>2} = \ldots$ The constraint states that the success probability in $R_k$ is larger than in $R_{>k}$, i.e., $p_k \geq p_{>k} := P(y=1 \,|\, R_{>k})$. A directional decision list that characterizes the negative class ($y=0$) is defined analogously as satisfying inequality (\ref{eq:condition}) with flipped sign. Directional decision lists are even more intuitive than general decision lists because the rules cannot alternate between explaining the two classes.

Learning of directional decision lists can be performed through greedy selection in the desired direction specified by the user. The only modification compared to the traditional version is that we enforce inequality (\ref{eq:condition}) at each step. Consequently, we relieve the algorithm from the responsibility to decide on the direction of the next rule. We call this a \emph{directed greedy search}. With this modification hill climbing indeed finds the global optimum in the tic-tac-toe example. If it is unknown which class is easier to characterize then we simply run the algorithm successively in both directions. This makes sure that the search is performed at least once in the easier direction.

\subsection{Monotone Probabilities}
A related family of decision lists was considered in \cite{Wang2015falling}. The constraint there is that the predicted probabilities corresponding to the ordered rules are monotonically decreasing (or increasing). This is an attractive property because it ensures that rules with the highest confidence appear at the top of the list. Decision lists with monotone probabilities are a strict subfamily of directional decision lists. In a directional decision list the `remainder probabilities' $p_{>k}$ are monotonically decreasing (or increasing) rather than the probabilities $p_k$ themselves. This may seem like a minor detail but it actually makes a big difference for the optimization. While directional decision lists can efficiently be learned with a greedy algorithm it is a much harder task to enforce monotone probabilities. Indeed, in our experiments hill climbing often failed to produce a decision list of the desired length satisfying this stronger monotonicity property. The reason for this is that given certain $k-1$ rules there simply may be no $k$-th rule which satisfies $p_{k-1} \geq p_k \geq p_{>k}$. However, there will always be a $k$-th rule that satisfies $p_k \geq p_{>k}$, which is the constraint we have to enforce for directional decision lists. In fact, the rule with the largest success probability achieves this. To learn decision lists with monotone probabilities the authors of \cite{Wang2015falling} propose to use simulated annealing, which is computationally much more expensive than a greedy search. For high-dimensional data they therefore perform a pre-mining step which selects a smaller number of features that are considered in the learning phase. With a small-width beam search it may (or may not, depending on the data) be possible to enforce monotone probabilities. However, our recommendation is to greedily learn a directional decision list, which in many cases simply happens to have monotone probabilities.

The experiments in \cite{Wang2015falling} show that for many datasets there exist accurate classifiers which satisfy the monotonicity property. This in particular suggests that our broader family of models will often be large enough to contain classifiers with competitive performance.

\subsection{Classification}
We now discuss the case where the only purpose of the directional decision list is classification, i.e., the corresponding probabilities are not of interest. The learning problem then boils down to finding a disjunction of conditions that approximate the desired class indicator. For example, if we would like to characterize the positive class then we are seeking binary expressions cond$_k$ ($k=1,\ldots,K$) such that
\[
\mathbbm{1}\{y=1\} \approx \mathbbm{1}\{\textrm{cond}_1 \, \vee \, \cdots \, \vee \, \textrm{cond}_K\}.
\]
The positive class is predicted if any of the $K$ conditions applies and the negative class is predicted otherwise. In particular, the order of the conditions does not matter anymore in this case. However, learning is still performed in a sequential way. Indeed, we can use the same directed greedy search as before and simply ignore the learned probabilities in the end. This approach provides an alternative to classical methods \cite{littlestone1988learning} for learning sparse disjunctions that works well even with noisy data. The type of classifier we obtain is also known as a cascading classifier and was used for example for face detection \cite{viola2004robust}. Our learning procedure is different though.

\begin{figure*}[!t]
\centering
\subfloat[]{\includegraphics[width=.45\textwidth]{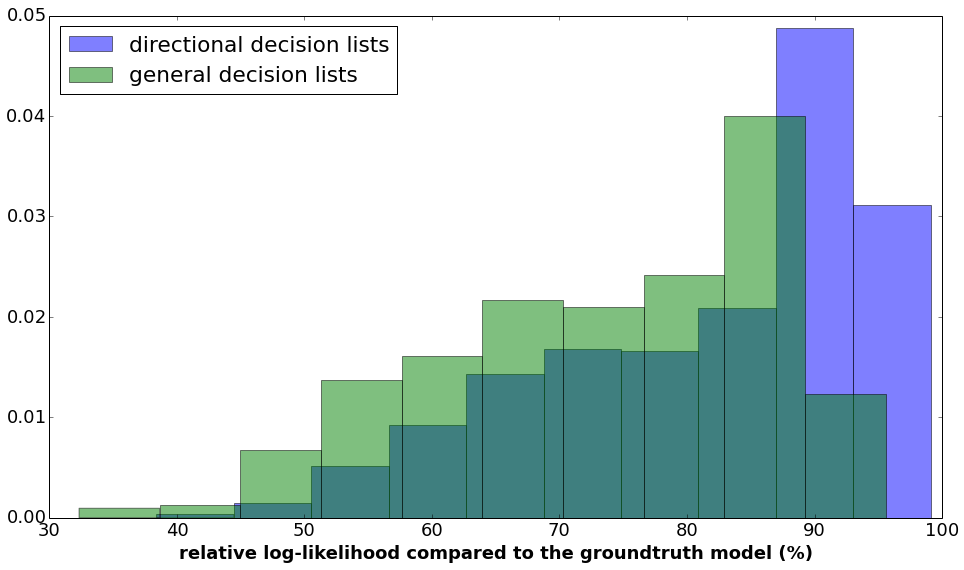}}
\hfil
\subfloat[]{\includegraphics[width=.45\textwidth]{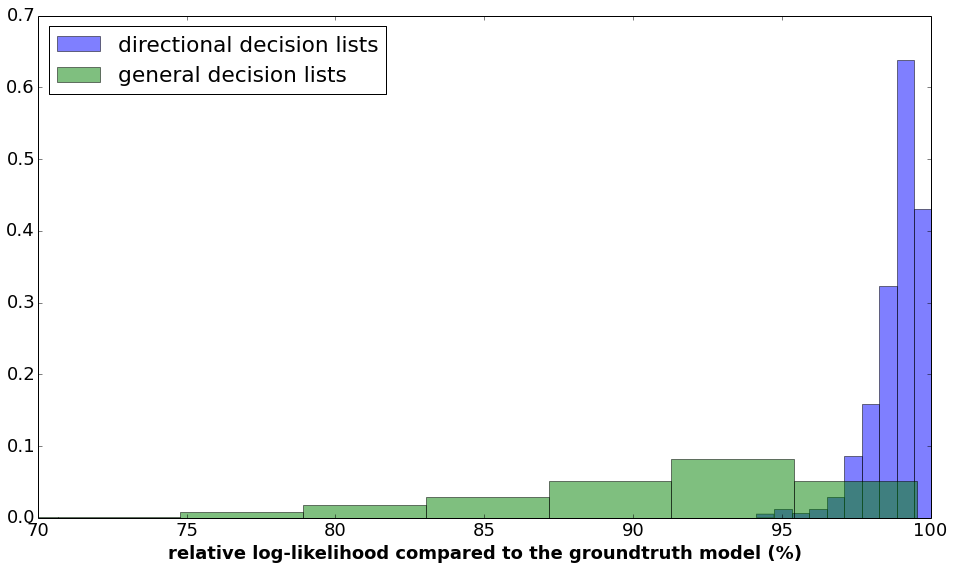}}
\caption{Histograms of likelihood performance in the synthetic experiment using (a) 1,000 and (b) 10,000 training examples.}
\label{fig:simulation}
\end{figure*}

\section{Experiments}
\label{sec:experiments}

\subsection{Simulated Data}
We use synthetic data to show that our proposed family of directional decision lists is more amenable to greedy learning than the unrestricted family of all decision lists. For that purpose we created random decision lists with 10 rules by choosing conditions from a pool of 1,000 binary features. The probabilities for general decision lists were simply drawn independently uniformly from [0,1]. To obtain directional decision lists we sorted the probabilities so that inequality (\ref{eq:condition}) holds. We then simulated training data by sampling binary features as well as corresponding labels according to the created ground-truth decision lists. Performance is evaluated by computing the ratio of the log-likelihood of the ground-truth model to the log-likelihood of the learned model. The likelihood is computed on an independent test set of 10,000 examples using random features and labels sampled from the corresponding ground-truth decision list. Figure \ref{fig:simulation} shows histograms over 1,000 simulation runs for both families of decision lists and different numbers of training examples. The performance of the greedy search for directional decision lists is significantly better than for the unrestricted family of all decision lists.

\subsection{Discriminative Features}
As a simple example of a classification task we consider the Titanic dataset\footnote{\url{http://www.amstat.org/publications/jse/v3n3/datasets.dawson.html}} which contains information about all Titanic passengers including labels indicating whether or not the person survived. We ran the hill-climbing algorithm to characterize the class of deceased passengers. The top three features we obtained were 1) male adult, 2) passenger in third-class, 3) adult in second-class. We also ran the algorithm in the other direction in order to find indicators for survival. The top three features we obtained were 1) female passenger, 2) child, 3) male passenger in first-class. Note that these results are different from simply ranking individual features based on their discriminative performance because additional features are chosen as to optimally complement the already selected features.

\subsection{Problem Symptoms in Manufacturing}
\begin{figure*}[!t]
\center
\includegraphics[width=.75\textwidth]{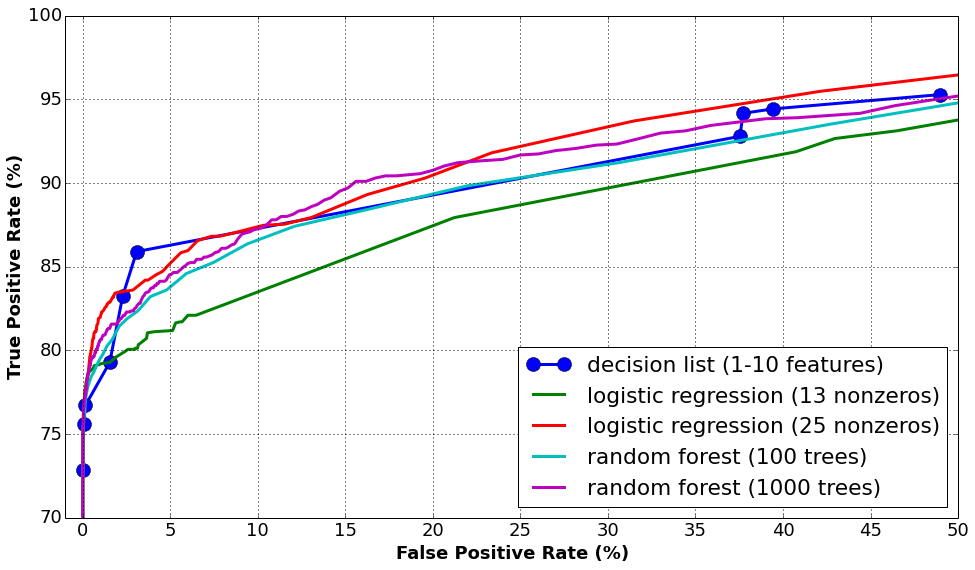}
\caption{ROC curves of different classifiers for the manufacturing data.}
\label{fig:roc}
\end{figure*}

We now demonstrate the practical usability of our approach through a manufacturing example. Interpretability is of central importance in manufacturing, especially when it comes to tasks like product quality control. In fact, the dataset we consider here motivated our work on interpetable classification. For some manufacturing applications interpretability is even a formal requirement. Our dataset consists of industrial process measurements for manufactured fuel pumps and the goal is to identify potential problem cases. The data originates from around 100,000 parts produced over a period of three months. We use the samples from the first two months for training and test on the samples from the third month. Each data point consists of around 400 attributes which are mainly numeric. We discretized the continuous variables based on empirical quantiles (\textless1\%, \textless5\%, \textless10\%, \textless25\%, $\in$[25\%,75\%], \textgreater75\%, \textgreater90\%, \textgreater95\%, \textgreater99\%). This yielded around 2,000 binary attributes. Each sample also has a label classifying it as a defective (NOK) or an intact (OK) part. This information comes from a detailed investigation of the parts. Around 96\% of the samples are labeled as OK while 4\% are NOK. The task is now to identify a small number of features which can discriminate between defective and intact products. This can be achieved in two ways. We could either try to characterize the class of intact products or try to characterize the class of defective products. It turned out that finding indicative features for the class of defective products is the easier direction. We used our algorithm to greedily learn a directional decision list of length up to 10. The selected features were mainly indicators for extreme values of the displaced volume under certain applied pressure levels. This is very useful information for the engineers because it directly points at the measurements which actually matter for determining the quality of the manufactured parts. As a consequence, test procedures can be improved. We show the true positive and false positive rates for the learned decision list in figure \ref{fig:roc}. Each dot corresponds to one additional rule. For comparison we trained state-of-the-art classifiers using the same discretized data. Sparse logistic regression models were learned using an L1-penalty with two different regularization strengths resulting in 13 and 25 nonzero parameters, respectively. We also learned random forests with 100 and 1,000 trees, respectively. By varying the threshold for classification we obtain ROC curves for each of these models, which are shown in figure \ref{fig:roc}. The performance of the decision list is comparable to the other models. However, the decision list has the advantage of being much more intuitive and providing very specific insight into the data. As a robustness check we ran a similar experiment on data from a second production line. The selected features were virtually the same as the ones chosen for the first production line. We also considered using pairs of primitive features as conditions in the decision list. However, this only led to a marginal improvement. For better interpretability we hence stayed with the primitive features. 

\section{Conclusion}
We proposed a restriction of the space of decision lists which is large enough to contain sufficiently accurate models and which allows for very efficient learning. This was achieved by introducing a definition for the direction of ordered probabilistic classification rules. Directional decision lists in particular generalize decision lists with monotone probabilities. In addition to providing essentially state-of-the-art classification performance in our experiment on real manufacturing data the introduced method successfully identified very meaningful discriminative features. While greedy algorithms are the traditional way to learn ordered rules, to our knowledge they have not been used before in conjunction with a directional constraint. Apart from being computationally efficient greedy algorithms have the additional advantage of being less susceptible to oversearching \cite{quinlan1995oversearching}. Indeed, in our experiments long-width beam searches tended to decrease interpretability and generalization performance. In that sense a greedy search can also be considered a form of regularization.



\bibliographystyle{IEEEtran}
\bibliography{references}

\end{document}